\documentclass[10pt,twocolumn,letterpaper]{article}

\usepackage{cvpr}
\usepackage{times}
\usepackage{epsfig}
\usepackage{graphicx}
\usepackage{amsmath}
\usepackage{amssymb}
\usepackage{booktabs}
\usepackage{subfig}

\usepackage[breaklinks=true,bookmarks=false]{hyperref}

\cvprfinalcopy 

\def\cvprPaperID{****} 

\begin{document}

\title{\emph{iCassava 2019} Fine-Grained Visual Categorization Challenge}

\author{Ernest Mwebaze, Timnit Gebru, Andrea Frome\\
Google Research\\
{\tt\small [emwebaze,tgebru, afrome]@google.com}
\and
Solomon Nsumba, Jeremy Tusubira\\
Artificial Intelligence lab\\
Makerere University\\
{\tt\small [snsumba, jtusubira]@gmail.com}
\and
Chris Omongo\\
National Crops Resources Research Institute\\
P.O. Box 7084 Kampala, Uganda.\\
{\tt\small chrisomongo@gmail.com}
}

\maketitle

\begin{abstract}

Viral diseases are major sources of poor yields for cassava, the 2$^{nd}$ largest provider of carbohydrates in Africa. At least 80\% of small-holder farmer households in Sub-Saharan Africa grow cassava. 
Since many of these farmers have smart phones, they can easily obtain photos of diseased and healthy cassava leaves in their farms, allowing the opportunity to use computer vision techniques to monitor the disease type and severity and increase yields. 
However, annotating these images is extremely difficult as experts who are able to distinguish between highly similar diseases need to be employed. 
We provide a dataset of labeled and unlabeled cassava leaves and formulate a Kaggle challenge to encourage participants to improve the performance of their algorithms using semi-supervised approaches. 
This paper describes our dataset and challenge which is part of the Fine-Grained Visual Categorization workshop at CVPR 2019.

\end{abstract}

\section{Introduction}


Cassava \emph{(Manihot esculenta Cranz)} is one of the key staples and food security crops in Africa because it is robust to adverse weather conditions and can be processed into multiple products \cite{11}. 
However, diseases that plague the crop are still a major challenge causing annual yield loss valued at an estimated 12-23 million dollars~\cite{otim-nape}. 
As the farmers do not know how to detect the disease, agricultural experts from the government have to visit farms to visually inspect the plants and determine whether they are are diseased or healthy. 
The labor intensive nature of this process and lack of many people who can reliably identify the disease, makes it difficult to monitor and treat disease progression.

Low cost methods of diagnosing cassava disease through computer vision algorithms running on farmers' smart phones could help monitor disease progression and increase yields. 

\begin{figure*}[t]
  \centering
      \subfloat[Healthy]{
      \includegraphics[width=.19\textwidth]{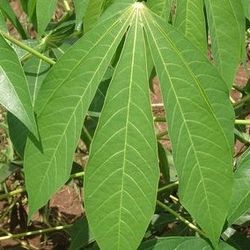}}
    \subfloat[CBB]{
      \includegraphics[width=.19\textwidth]{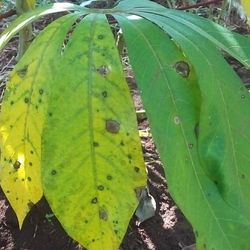}}
    \subfloat[CGM]{
      \includegraphics[width=.19\textwidth]{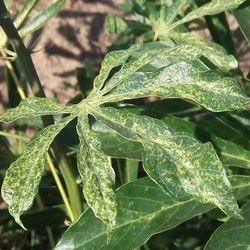}}
    \subfloat[CMD]{
      \includegraphics[width=.19\textwidth]{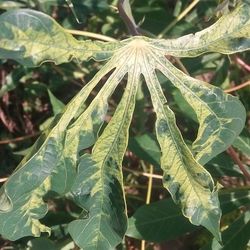}}
    \subfloat[CBSD]{
      \includegraphics[width=.19\textwidth]{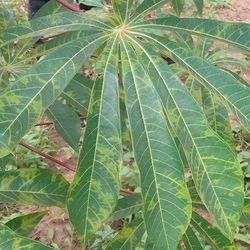}}
    \caption{Prototypical images associated with the five classes in our dataset covering healthy cassava leaves as well as $4$ common diseases.}
    \label{sampleimages}
\end{figure*}

To encourage the development of such tools, we present a dataset of cassava leaves, as well as a Kaggle\footnote{https://www.kaggle.com/c/cassava-disease} challenge focused on accurately distinguishing between four of the most common cassava diseases: Cassava Brown Streak Disease (CBSD), Cassava Mosaic Disease (CMD), Cassava Bacterial Blight (CBB) and Cassava Green Mite (CGM). Figure \ref{sampleimages} presents distinct prototypical images of the diseases. Of the four, CMD and CBSD are the most devastating diseases to the cassava yield in Eastern and Central Africa \cite{15,16} and the greatest threats to the food security and livelihoods of over $200$ million people.

Cassava disease can only be identified by trained experts as the different types of disease show very similar symptoms, and each leaf can have multiple disease further complicating the task. Thus, gathering labeled training data is labor intensive and expensive. To account for this real world limitation, our challenge contains unlabeled data to encourage participants to work on algorithms that utilize the more abundantly available unlabeled training data, as well as labeled data annotated by experts.

\subsection{Problem Context}

To diagnose cassava disease, experts from government agencies travel to disparate parts of the country and visually score the plants by investigating the leaves for disease symptoms.
This method tends to be very subjective; it is not uncommon for experts to disagree on a diagnosis for a particular plant. 
An automated tool could potentially guide experts to diagnose disease more reliably and enable farmers in remote places to monitor their crops without experts. 

There are multiple factors which complicate the design of such a tool.

(1) Images used by the system will by design be taken \emph{in-situ}, in the farmers’ fields by the farmers. So the solution should be robust towards different viewpoints, backgrounds and varying amounts of noise in the input images. 
(2) Disease symptoms can be very similar and a single leaf can have multiple diseases with varying levels of  severity.  
(3) Due to the difficulty of annotating cassava disease, there are many more unlabeled images than those annotated by experts. Thus, algorithms that utilize an abundance of unlabeled images and minimal labeled ones are preferred.
(4) The solution should be able to run on the farmers’ phones, requiring a fast and light-weight model with minimal access to the cloud.

In this challenge, we encouraged participants to address the first $3$ factors in their solutions.
Our dataset was crowdsourced from approximately $200$ farmers with farms located in different parts of Uganda. 
While the pictures were taken with the same type of phone, images from each farm can have different characteristics. We screened these images for obvious data errors such as having photos of objects that are not plants, but tried to minimize the amount of data cleaning in order to have a dataset that reflects the context under which an automated cassava disease diagnosis tool would be deployed. 
To simplify the challenge, we only labeled each leaf with the most significant disease identified by experts from the government agency in charge of cassava disease surveillance.
A follow up challenge will have a dataset that reflects multiple disease labels per leaf along with the severity of each label, and a setup that addresses the $4^{th}$ necessity of a light-weight model.

\subsection{Related work}

Machine learning for agriculture is a nascent field with little prior research pertaining to cassava as well as other crops~\cite{1,6,7,8}. 
Due to the difficulty of annotating crops for automated disease diagnosis, most of the algorithms used to monitor plants need to be able to achieve good performance with small samples. Many prior works aim to solve a binary classification problem: distinguishing diseased plants from healthy ones. 
They also acquire images in controlled environments whereas diagnostic tools deployed in different farms need to be robust to many types of variations (e.g. lighting, background, terrain).

Due to the success of deep learning based methods, many works now infer plant diseases from images using convolutional neural networks~\cite{neuNet1, neuNet2}. 
While this approach automates the feature extraction process that was a major bottleneck in previous methods, the amount of data and increased processing time required at training and prediction time means that they are slow to deploy on low resource gadgets in the field. While many other digital image processing techniques have been used in the literature, we will not discuss them here for brevity but point readers to~\cite{review1} for a literature review.

\section{The Cassava disease leaf Data}
Our dataset consists of $9,436$ labeled and $12,595$ unlabeled images of cassava plant leaves. The annotations consist of $5$ classes; healthy plant leaves (316/211 train/test examples) and diseased plant leaves representing the 4 diseases; CMD (2658/1773 train/test images), CBSD (1443/963 train/test images), CBB (466/311 train/test images), and CGM (773/516 train/test images). Figure \ref{sampleimages} depicts typical leaf images for each of the $4$ types of diseases.  

The data was collected as part of a crowdsourcing project by the Artificial Intelligence lab\footnote{www.air.ug} in Makerere University and the National Crops Resources Research Institute, (NaCRRI)\footnote{www.naro.go.ug} using smart phones. NaCRRI is the Ugandan governmental body responsible for agricultural research in the country. All the collected images were manually labelled by experts from NaCRRI who scored each of the images for disease incidence and severity. For this challenge however, we only employ the disease class annotations (without severity).

\subsection{Disease leaf symptoms}
Each of the $4$ cassava diseases included in our dataset causes unique symptomatic features to appear on the leaves as shown in Figure \ref{sampleimages}. Although images in Figure~\ref{sampleimages} show distinctive prototypes to clearly highlight each disease’s symptoms, data from the field appears more blended with multiple diseases inflicting a single plant. The four major diseases affecting cassava and their major symptoms include:

\subsubsection{Cassava mosaic disease (CMD)} The most widespread cassava disease in sub-Saharan Africa, CMD produces a variety of foliar symptoms that include mosaic, mottling, misshapen and twisted leaflets, and an overall reduction in size of leaves and plants \cite{3}. Leaves affected by this disease have patches of normal green color mixed with different proportions of yellow and white depending on the severity. These chlorotic patches indicate reduced amounts of chlorophyll in the leaves, which affects photosynthesis and thus limits crop yield. 

\subsubsection{Cassava brown streak disease (CBSD)} CBSD is presently the most severe of the cassava diseases. It is vectored by white flies and can also be transmitted through infected cuttings. The disease is very common in East Africa and in other cassava growing countries in sub-Saharan Africa. CBSD leaf symptoms consist of a characteristic yellow or necrotic vein banding which may enlarge and coalesce to form comparatively large yellow patches. Tuberous root symptoms consist of dark-brown necrotic areas within the tuber and reduction in root size and according to \cite{2}, leaf and/or stem symptoms can occur without the development of tuber symptoms. 

\subsubsection{Cassava bacterial blight (CBB)} CBB is a major bacterial disease which is common in moist areas. The predominance and severity of its symptoms can vary depending on location, season and aggressiveness of the bacterial strains. CBB leaf symptoms include; black leaf spots and blights, angular leaf spots, and premature drying and shedding of leaves due to the wilting of young leaves and severe attack.

\subsubsection{Cassava green mite (CGM)}
This disease causes white spotting of leaves, which increase from the initial small spots to cover the entire leaf causing loss of chlorophyll. Leaves damaged by CGM may also show mottled symptoms which can be confused with symptoms of cassava mosaic disease (CMD). Severely damaged leaves shrink, dry out and fall off, which can cause a characteristic candle-stick appearance. 

\subsection{Contextual complexities of the dataset}
\begin{figure}[t]
  \centering
      \subfloat[Background effects]{
      \includegraphics[width=.24\textwidth]{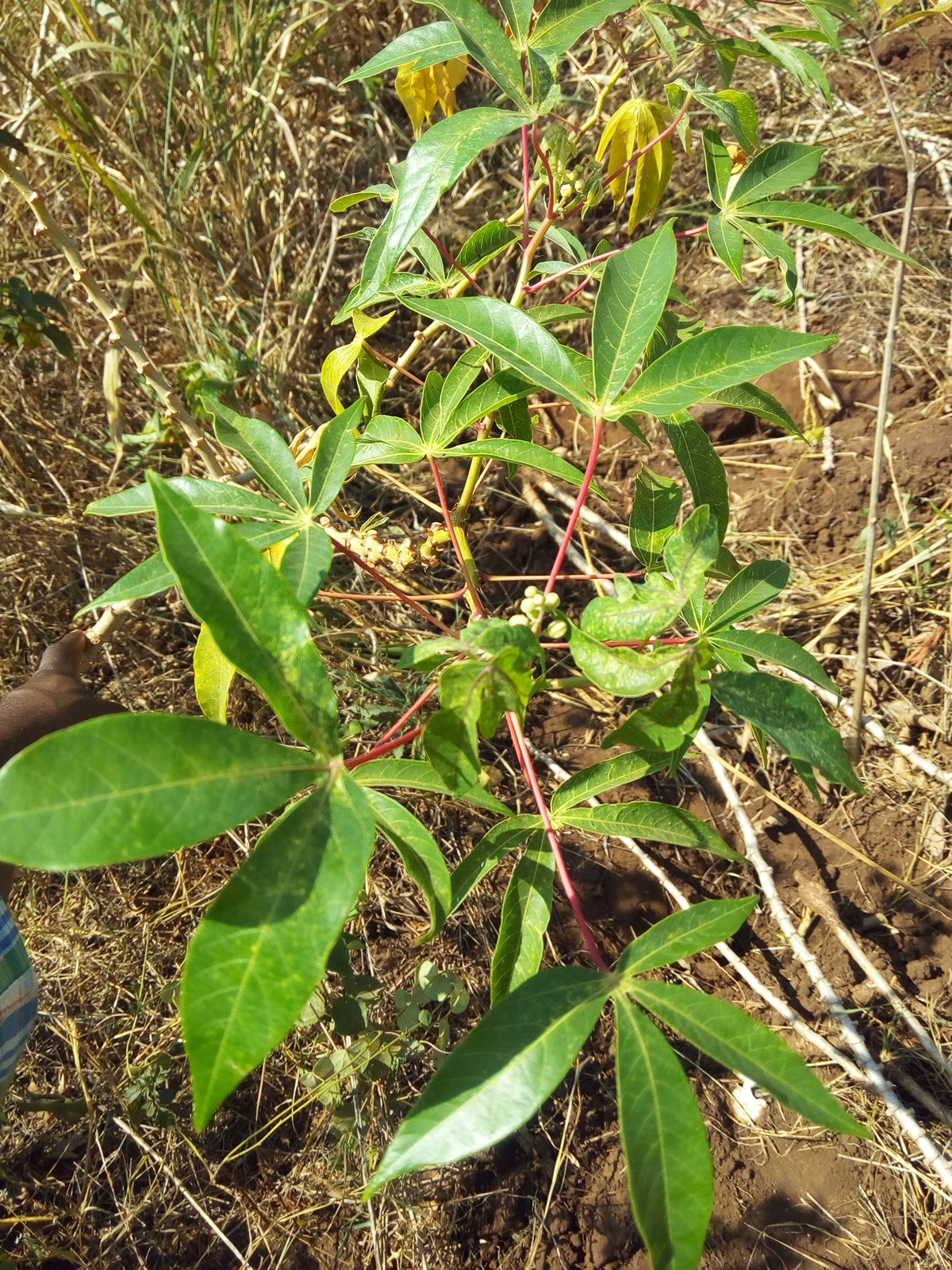}}
    \subfloat[Time of day effects]{
      \includegraphics[width=.24\textwidth]{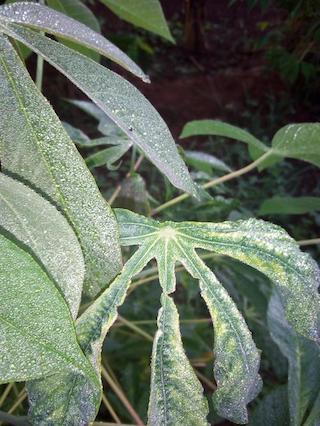}}
            \\
    \subfloat[Multiple disease]{
      \includegraphics[width=.24\textwidth]{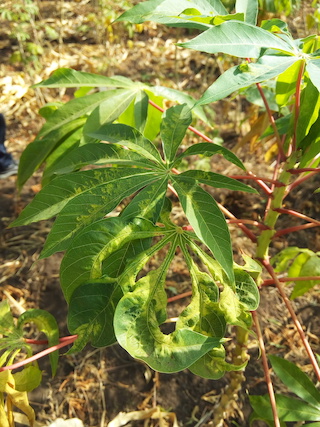}}
    \subfloat[Poor focus effects]{
      \includegraphics[width=.24\textwidth]{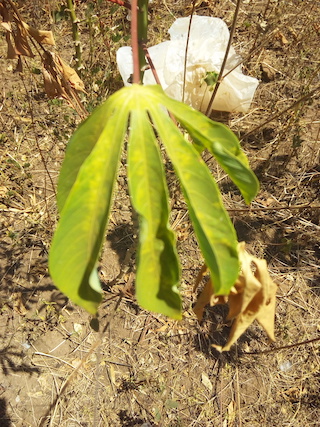}}
    \caption{Examples complexities in the leaf images used in the challenge.}
    \label{fig:imagecomplexities}
\end{figure}

We identify $4$ factors that may increase the difficulty of classifying cassava disease in our dataset: (1) the different image backgrounds and scales, (2) the time of day the image was acquired, (3) multiple co-occurring diseases on one plant and (4) the poor focus of some images. Figure \ref{fig:imagecomplexities} gives an example of each of these issues in the dataset. 

Many of these factors are difficult to avoid. For example, Figure~\ref{fig:imagecomplexities}a represents a typical background of an image taken in a smallholder farmer’s garden. Because there are many farmers, the images are acquired from multiple gardens making it impossible to have a uniform background in the field. While there have been prior attempts to provide a background sheet for the farmers to take images against, this method proved to be too cumbersome and was eventually abandoned. 

Figure~\ref{fig:imagecomplexities}b was taken early in the morning which is when most farmers will go to their gardens. As seen in the figure, the leaf is covered in dew adding to the complexity of identifying the disease from such an image. This challenge is compounded by the co-occurrence of multiple diseases on the same plant as shown in Figure~\ref{fig:imagecomplexities}c. Since most cassava varieties have varying tolerance levels for the different diseases, one plant can have multiple diseases with varying securities making it difficult to diagnose. In addition to these complexities, as shown in Figure~\ref{fig:imagecomplexities}d, the images are not always in focus even after farmers have been given instructions to center images on the leaf of interest. 

\section{iCassava challenge Data preparation}
One of the goals of this challenge was to expose the computer vision community to a realistic dataset that represents existing challenges faced by many people in the world. Thus, while processing the data, we were careful to ensure that the dataset still contained elements of the contextual challenges described above. The data preparation pipeline involved four main steps.

\begin{enumerate}
  \item Data acquisition
  Together with the National Crops Resources Research Institute (NaCRRI) in Uganda, we developed and deployed a crowdsourcing system where small-holder farmers in disparate places in Uganda were given smart phones with an application used to collect images of the crops in the farmers' fields. Data was collected over a period of 1 year with the goal of estimating the health of cassava gardens in Uganda. Approximately 200 farmers sent images of plants from their gardens over the course of 1 year. 
  \item Data cropping
  As described above, images collected \emph{in situ} have varied backgrounds and scales because they were taken by various farmers at different times over a period of 1 year. We additionally processed some of the images by cropping them in order to have the cassava leaves in the center of the image frame. 
  \item Data annotation
  The leaf images were annotated in collaboration with NaCRRI’s cassava diagnostics domain experts. The experts annotated each leaf image with the primary cassava disease as well as the secondary one if it was evidently displayed in the image. They also scored the disease severity on a scale of 1 to 5, with 1 representing a healthy leaf and 5 a severely diseased one. For this challenge however we only provided the primary disease label associated with each image. 
 \item Data verification
  The final task in our data processing pipeline involved consistency checks such as removing duplicate images and re-annotating or removing images with inconsistent labels. Inconsistent labels arose from more than one expert annotating the same image differently. Because an image can have more than one leaf with different diseases per leaf, this is possible albeit challenging to decide which label to use. For cases like these, we dis-included the images in the challenge dataset.

\end{enumerate}

\section{Submitted challenge solutions}
\begin{figure*}
\centering
\includegraphics[width=0.9\textwidth]{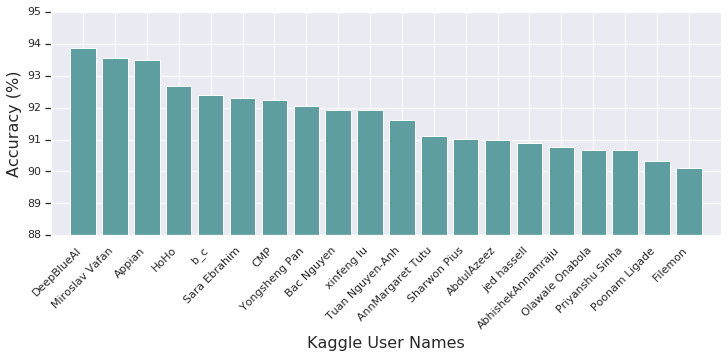} 
 \caption{Cassava disease classification final challenge results.}
 \label{fig:challenge_scores}
\end{figure*}

The cassava disease classification challenge ran on Kaggle from April 25 to June $1^{st}$ and garnered 87 participants. The evaluation metric was overall accuracy, with the top $3$ contenders attaining a score of $\sim{93\%}$ and the lowest scoring $3$ participants achieving an accuracy of $\sim{19\%}$. The challenge had two modes of evaluation; the first being the public leaderboard where the user scores were calculated on an unknown 40\% of the testset, and the second a private leaderboard where candidates were scored on the entire testset. Figure~\ref{fig:challenge_scores} shows the final challenge accuracy of the top 20 teams from the private leaderboard. The winning entry was able to use the unlabeled images as additional training data to slightly improve accuracy (by $\sim{1\%}$). All $3$ solutions used the ResNet~\cite{resnet} architecture with some data augmentation.

\section{Discussion}
We presented a dataset of cassava plants annotated with the types of disease inflicting the plants, as well as an unlabeled set of cassava images taken by farmers. Our dataset and accompanying Kaggle challenge hope to encourage the computer vision community to work on problems that affect the very serious issue of food security. Fine-grained image classification can be used to help small-holder farmers monitor their plants and increase yields. 

While the research community has made significant improvements in fine-grained image classification using large training data, as can be seen in our setting, many real world problems require the use of smaller datasets and semi-supervised methods to leverage unlabeled images. We plan to release additional datasets and challenges which incorporate the extra complexities arising from multiple diseases associated with each plant as well as varying levels of severity. 

\section*{Acknowledgment}
The authors gratefully acknowledge the team at National Crops Resources Research Institute (NaCRRI) for their unconditional assistance especially in the data collection and annotation stages. The authors would also like to acknowledge the Bill and Melinda Gates Foundation and USAID who provided the requisite funding for the project that supported this data collection.

\bibliographystyle{IEEEtran}
\bibliography{icassavacvpr}

\begin{thebibliography}{10}
\providecommand{\url}[1]{#1}
\csname url@samestyle\endcsname
\providecommand{\newblock}{\relax}
\providecommand{\bibinfo}[2]{#2}
\providecommand{\BIBentrySTDinterwordspacing}{\spaceskip=0pt\relax}
\providecommand{\BIBentryALTinterwordstretchfactor}{4}
\providecommand{\BIBentryALTinterwordspacing}{\spaceskip=\fontdimen2\font plus
\BIBentryALTinterwordstretchfactor\fontdimen3\font minus
  \fontdimen4\font\relax}
\providecommand{\BIBforeignlanguage}[2]{{%
\expandafter\ifx\csname l@#1\endcsname\relax
\typeout{** WARNING: IEEEtran.bst: No hyphenation pattern has been}%
\typeout{** loaded for the language `#1'. Using the pattern for}%
\typeout{** the default language instead.}%
\else
\language=\csname l@#1\endcsname
\fi
#2}}
\providecommand{\BIBdecl}{\relax}
\BIBdecl

\bibitem{11}
C.~Poulton, G.~Tyler, P.~Hazell, A.~Dorward, J.~Kydd, and M.~Stockbridge,
  ``Competitive commercial agriculture in sub?saharan africa,'' \emph{Centre
  for Environmental Policy, Imperial College London, Wye, Ashford, Kent, TN25
  5AH,UK}, 2006.

\bibitem{otim-nape}
G.~W. OTIM-NAPE, T.~ALICAI, and J.~M. THRESH, ``Changes in the incidence and
  severity of cassava mosaic virus disease, varietal diversity and cassava
  production in uganda,'' \emph{Annals of Applied Biology}, vol. 138, no.~3,
  pp. 313--327, 2001.

\bibitem{15}
E.~Nuwamanya, Y.~Baguma, E.~Atwijukire, S.~Acheng, and T.~Alicai, ``Competitive
  commercial agriculture in sub?saharan africa,'' \emph{International Journal
  of Plant Physiology and Biochemistry}, vol. 7(2), pp. 12--22, 2015.

\bibitem{16}
G.~M. Rwegasira and C.~M.~E. Rey, ``Response of selected cassava varieties to
  the incidence and severity of cassava brown streak disease in tanzania,''
  \emph{Journal of Agricultural Science}, vol.~4, no.~7, 2012.

\bibitem{1}
E.~Mwebaze and M.~Biehl, ``Prototype-based classification for image analysis
  and its application to crop disease diagnosis,'' \emph{Advances in
  Self-Organizing Maps and Learning Vector Quantization - Proceedings of the
  11th International Workshop {WSOM} 2016}, pp. 329--339, January 2016.

\bibitem{6}
J.~R. Aduwo, E.~Mwebaze, and J.~A. Quinn, ``Automated vision-based diagnosis of
  cassava mosaic disease,'' \emph{Industrial Conference on Data Mining}, pp.
  114--122, 2010.

\bibitem{7}
E.~Mwebaze, P.~Schneider, F.~Schleif, J.~R. Aduwo, J.~A. Quinn, S.~Haase,
  T.~Villmann, and M.~Biehl, ``Divergence-based classification in learning
  vector quantization,'' \emph{Neurocomputing}, vol.~74, pp. 1429--1435, 2011.

\bibitem{8}
J.~Tuhaise, J.~A. Quinn, and E.~Mwebaze, ``Pixel classification methods for
  automatic symptom measurement of cassava brown streak disease,''
  \emph{Proceding of the 1st International Conference on the Use of Mobile ICT
  in Africa}, 2014.

\bibitem{neuNet1}
S.~Sladojevic, M.~Arsenovic, A.~Anderla, D.~Culibrk, , and D.~Stefanovic,
  ``{Deep Neural Networks Based Recognition of Plant Diseases by Leaf Image
  Classification},'' \emph{{Computational Intelligence and Neuroscience}}, vol.
  2016, p.~11, 2016.

\bibitem{neuNet2}
S.~P. Mohanty, D.~P. Hughes, and M.~Salath{\'{e}}, ``Using deep learning for
  image-based plant disease detection,'' \emph{CoRR}, vol. abs/1604.03169,
  2016.

\bibitem{review1}
A.~Barbedo and J.~Garcia, ``Digital image processing techniques for detecting,
  quantifying and classifying plant diseases,'' \emph{SpringerPlus}, vol.~2,
  no.~1, pp. 1--12, 2013.

\bibitem{3}
I.~Abdullahi, G.~Atiri, and A.~Dixon, ``Effects of cassava genotype, climate
  and the bemisia tabaci vector population on the development of african
  cassava mosaic geminivirus (acmv),'' \emph{Acta Agronomica Hungarica}, pp.
  285--289, 2003.

\bibitem{2}
R.~Hillocks, M.~Raya, and J.~Thresh, ``The association between root necrosis
  and above-ground symptoms of brown streak virus infection of cassava in
  southern tanzania,'' \emph{International Journal of Pest Management}, pp.
  285--289, 1996.

\bibitem{resnet}
K.~He, X.~Zhang, S.~Ren, and J.~Sun, ``Deep residual learning for image
  recognition,'' in \emph{Proceedings of the IEEE conference on computer vision
  and pattern recognition}, 2016, pp. 770--778.

\end{thebibliography}
\end{document}